\documentclass[hidelinks]{article}
\usepackage[preprint]{neurips_2022}
\usepackage[utf8]{inputenc} 
\usepackage[T1]{fontenc}    
\usepackage{hyperref}       
\usepackage{url}            
\usepackage{booktabs}       
\usepackage{amsfonts}       
\usepackage{nicefrac}       
\usepackage{microtype}      
\usepackage{xcolor}         
\usepackage{algorithm}
\usepackage{algorithmic}
\usepackage{float}
\usepackage{wrapfig}
\usepackage[utf8]{inputenc}
\usepackage[small]{caption}
\usepackage{algorithm}
\usepackage{natbib}
\usepackage{algorithmic}
\usepackage{graphicx}
\usepackage{subfigure}
\usepackage{amsmath}
\usepackage{dsfont}
\usepackage{booktabs}
\usepackage{tabularx}
\usepackage{booktabs}
\usepackage{comment}
{\newtheorem{assumption}{Assumption}}
\usepackage{amsmath}
{}
\usepackage{amsmath}
{\newtheorem{lemma}{Lemma}}
\usepackage{amsmath}
{\newtheorem{corollary}{Corollary}}
\usepackage{amsmath}
{\newtheorem{theorem}{Theorem}}
\usepackage{amsmath}
{\newtheorem{remark}{Remark}}

\DeclareMathOperator*{\argmax}{arg\,max}
\DeclareMathOperator*{\argmin}{arg\,min}
\newcommand{\pscal}[2]{\left\langle #1, #2 \right\rangle}

\newcommand{\Abs}[1]{\left\lVert #1 \right\rVert}

\title{Federated Learning Aggregation: New Robust Algorithms with Guarantees}
\author{%
  Adnan Ben Mansour\\
  Department of Computer Science\\
  ENS, PSL Research University\\
  Paris, France\\
  \texttt{adnan.ben.mansour@ens.psl.eu} \\
  \And
  Gaia Carenini\\
  Department of Computer Science\\
  ENS, PSL Research University\\
  Paris, France\\
  \texttt{gaia.carenini@ens.psl.eu} \\
  \AND 
  Alexandre Duplessis\\
  Department of Computer Science\\
  ENS, PSL Research University\\
  Paris, France\\
  \texttt{alexandre.duplessis@ens.psl.eu} \\
  \And
  David Naccache\\
  Department of Computer Science\\
  ENS, CNRS, PSL Research University\\
  Paris, France\\
  \texttt{david.naccache@ens.fr} \\
}
\begin{document}
\maketitle
\begin{abstract}
Federated Learning has been recently proposed for distributed model training at the edge. The principle of this approach is to aggregate models learned on distributed clients to obtain a new more general ``average'' model (FedAvg). The resulting model is then redistributed to clients for further training. To date, the most popular federated learning algorithm uses coordinate-wise averaging of the model parameters for aggregation. In this paper, we carry out a complete general mathematical convergence analysis to evaluate aggregation strategies in a federated learning framework. From this, we derive novel aggregation algorithms which are able to modify their model architecture by differentiating client contributions according to the value of their losses. Moreover, we go beyond the assumptions introduced in theory, by evaluating the performance of these strategies and by comparing them with the one of FedAvg in classification tasks in both the IID and the Non-IID framework without additional hypothesis.
\end{abstract}
\section{Introduction}
Until recently, machine learning models were extensively trained in a data center setting using powerful computing nodes, fast inter-node communication links, and large centrally available training datasets. The future of machine learning lies in moving both data collection as well as model training to the edge. 
Moreover, nowadays, more and more collected data-sets are privacy sensitive, in particular in confidentiality-critical fields such as patients' medical information processing. Specifically, reducing human exposure to data is highly desirable to avoid confidentiality violations due to human failure. This may preclude logging to a data center and performing training there using conventional approaches. Indeed, conventional machine learning requires feeding training data into a learning algorithm and revealing information (albeit indirectly) to their developers; moreover, when several data sources are involved, a merging procedure for creating a single dataset is also required. In 2017, \cite{McMahan2017CommunicationEfficientLO} advocated an alternative distributed data-mining technique for edge devices, termed \emph{Federated Learning} (FL), allowing a decoupling of model training from the need for direct access to the raw training data. 
\paragraph{Federated learning protocol}
Formally, FL is a protocol acting as per Algorithm \ref{alg:generic_FL}, (cf. \cite{Li2020FederatedLC} for an overview). The framework involves a group of devices named \emph{clients} and a \emph{server} coordinating the learning process. Each client has a local training dataset which is never uploaded to the server. The goal is to train a global model by aggregating the results of the local training clients.
Parameters, fixed by the centralized part of the global learning system, include: a set $I$ grouping \emph{N} clients, the ratio of clients \emph{C} selected at each round, the number of communication rounds \emph{T} and a number of local epochs \emph{E}. The model is defined by its weights : at the end of each epoch $t \in \{ 0, \dots, TE -1 \}$, $w_{t+1}^i$ defines the weight of client $i \in I$. For each communication round $t \in \{ 0, E, \dots, (T-1)E \}$, $w_t$ is the global model detained by the server and $w_{TE}$ is the final weight.
\begin{algorithm}[tb]
\caption{Generic Federated Learning Algorithm}
\label{alg:generic_FL}
\textbf{Input}: \emph{N}, \emph{C}, \emph{T}, \emph{E} \\
\textbf{Output}: $w_{TE}$
\begin{algorithmic}[1] 
\STATE Initialize $w_0$.
\FOR{each round $t\in\{0,E, 2E,\dots,(T-1)E\}$}
\STATE $m \leftarrow \max(C \cdot N, 1)$
\STATE $I_t \leftarrow (\text{random set of \emph{m} clients})$
\FOR{each client $i \in I_t$ \textbf{in parallel}}
\STATE $w_{t+E}^{i} \leftarrow \text{\textsc{Client-Update}}(w_t)$
\ENDFOR
\STATE $w_{t+E} \leftarrow \text{\textsc{Aggregation}}(w_{t+E}^1, \dots, w_{t+E}^N)$
\ENDFOR
\STATE \textbf{return} $w_{TE}$
\end{algorithmic}
\end{algorithm}
 Algorithm \ref{alg:generic_FL} encodes the training procedure described below. There is a fixed set of $I = \{1,\dots,N\}$ clients (each with a local dataset), before every communication round $t \in \left\{0, E, \dots, (T-1)E \right\}$ the server randomly selects a set $I_t$ of $C\cdot N$ clients, the server sends to the clients the current global algorithm state, then the server asks the clients to perform local computations based on the global state and their local dataset and to send back an update; at the end, the server updates the weights of the model by aggregating clients' updates and the process repeats. For sake of generality, no structure is specified for the local training procedure (\emph{Client-Update}): give that different methods can be employed, for instance mini-batch SGD (\cite{Gower2019SGDGA}), Newton methods (\cite{Berahas2019LimitedmemoryBW}) or PAGE (\cite{DBLP:journals/corr/abs-2108-04755}). Similarly, no function for model aggregation is given yet. 
\paragraph{Model aggregation} 
To date, several aggregating functions have been proposed to accomplish this task. In 2017, the seminal work of \cite{McMahan2017CommunicationEfficientLO} proposed a plain coordinate-wise mean averaging of model weights. In 2018, \cite{ek:hal-03207411} adjusted this to enforce closeness of local and global updates. In 2019, \cite{Yurochkin2019BayesianNF} proposed an extension taking the invariance of network weights to permutation into account. More recently, in 2020, \cite{Cho2020ClientSI} extended the  coordinate-wise mean averaging substituting it by a term amplifying the contribution of the most informative terms against less informative one. Despite methodological advances, there is neither theoretical nor practical evidence for the right criterion for choosing a particular aggregation strategy.
\paragraph{Our contribution}  In this paper, we develop an extended mathematical analysis of aggregation strategies for FL, inspired by the article of \cite{Cho2020ClientSI}. We specifically focus on the trade-off among accuracy and convergence speed in the aggregation step of FL protocol. We propose several strategies either derived using the theoretical framework built (FedWorse, FedSoftWorse) or inspired by analogy (FedWorse($k$)) or opposition (FedBetter, FedSoftBetter, FedBetter($k$)). We study  extensively the properties of each algorithm  in classification task (both in an IID and a non IID framework); in this empirical analysis, we ignore consciously the assumptions adopted in theory with the goal of investigating their possible relaxation. Furthermore, we introduce a class of hybrid methods combining FedAvg and the novel aggregations proposed for maximizing efficiency and stability. 
\section{Analysis of aggregation algorithms for Federated Learning}
In this section, we study mathematically the convergence of federated learning aggregation methods under reasonable assumptions \ref{assumption}. Moreover, we discuss some algorithmic consequences of the theoretical result obtained.
\subsection{Framework of the analysis} \label{framework}
From now on, we will assume that all the clients are involved in each local and global iteration ($\forall t, I_t=I$); moreover, we will denote by $F_i$ the loss function of the $i$-th client and by $F$ the weighted average of the $F_i$ upon the distribution $P:=\{p_i\mid i\in I\}$. We restrain our analysis to the case in which \textsc{Client-Update} is mini-batch SGD with learning rate decay $\eta_t$ and mini-batch $\zeta_t^i$. In particular, we define
$g_i(w_t^i)$ as $\frac{1}{b}\sum\limits_{\zeta\in \zeta_t^i} \nabla F_i(w_t^{i},\zeta)$, where $b$ is the cardinality of the mini-batch.\\
For any iteration, we compute the weight as below: 
\begin{equation}
   w_{t+1}^i := \left\{
    \begin{array}{ll}
        \displaystyle w_t^i - \eta_t g_i(w_t^i) & \mbox{if } t \neq 0 \bmod E \\
        \\
        \displaystyle\sum_{j\in I} \alpha_t^j (w_t^j - \eta_t g_j(w_t^j)) := w_{t+1} & \mbox{if } t = 0 \displaystyle\bmod E
    \end{array} \right.
\end{equation}
where $\alpha_t^j$ is the aggregation coefficient referred to client $j$ at communication round $t$, with $\sum\limits_{j \in I}{\alpha_t^j} = 1$ for each $t$.
\subsection{Assumptions of the analysis}\label{assumption}
We enumerate here the assumptions under which we develop our model: 
\begin{assumption}{\textbf{\textit{$L$-smoothness}}}\label{assumption1}
$F_1,\dots,F_N$ satisfy:\\$\forall v,w$, $F_{i}(v)\leq F_i(w)+\pscal{v-w}{\nabla F_i(w)}+\frac{L}{2}\Abs{v-w}_2^2$.
\end{assumption}
\begin{assumption}{\textbf{\textit{$\mu$-convexity}}}\label{assumption2}
$F_1,\dots,F_N$ satisfy:\\ $\forall v,w$, $F_{i}(v)\geq F_i(w)+\pscal{ v-w}{\nabla F_i(w)} +\frac{\mu}{2}\Abs{v-w}_2^2$.
\end{assumption}
\begin{assumption}{\textbf{\textit{Bounded gradient estimator}}}\label{assumption3}
$\forall i\in I$, $\mathds{E}\Abs{g_i(w_i)-\nabla F_i(w_i)}^2\leq \sigma^2$.
\end{assumption}
\begin{assumption}{\textbf{\textit{Uniformly bounded expected squared norm}}}\label{assumption4}
$\forall i\in I$, $\mathds{E}\Abs{g_i(w_i)}^2\leq G^2$.
\end{assumption}
\begin{remark}
    Assumption \ref{assumption3} and Assumption \ref{assumption4} can be merged into a unique assumption of the form $\mathds{E}\Abs{\nabla F_i(w_i)}^2\leq A^2$.
\end{remark}
\subsection{Preliminaries}
The following is inspired by \cite{Cho2020ClientSI}. However, we extend it to the more general framework where instead of analyzing the choice of the clients, we look at the weights' attribution. Moreover we go beyond the work of \cite{Cho2020ClientSI} by analyzing the expression of the learning error in order to find out optimal strategies.
To facilitate the convergence analysis, we define the quantity $w_t$ for $t\neq 0 \bmod E$ as:
\begin{equation}
    w_{t+1}:=w_t - \eta_t\sum\limits_{i\in I} \alpha_t^i g_i(w_t^i)
\end{equation}
where $\alpha_t^i=p_i$.
Let $w^\star$ be the global optimum of $F$ and $w^\star_i$ the global optimum of $F_i$. We  define $F^\star$ as $F(w^\star)$, $F^\star_i$ as $F(w^\star_i)$ and the \emph{heterogeneity} as:
\begin{equation}
    \Gamma:=F^\star-\sum_{i\in I} p_i F_i^\star
\end{equation}
We now state two useful results the proof of which are contained in subsection \ref{Proof preliminary Lemmas} of the appendix:
\begin{lemma}\label{smoothRes}
Let $f$ be a $L$-smooth function with a unique global minimum at $w^\star$. Then :
\begin{equation}
    \forall w, ||\nabla f(w)||^2\leq 2L(f(w)-f(w^\star))
\end{equation}
\end{lemma}
\begin{lemma}\label{discrepency}
With the same notations as above and defining $\mathds{E}[.]$ as the total expectation over all random sources, the expected average discrepancy between $w_t$ and $w_t^i$ is bounded:
\begin{equation}
   \displaystyle\mathds{E} \left[\sum_{i\in I} \alpha_t^i \Abs{w_t-w_t^i}^2\right] \le 16\eta_t^2 E^2 G^2
\end{equation}
\end{lemma}
Before presenting the main results, we define the \emph{weighting skew} $\rho$ as: 
\begin{equation}\label{rho}
    \rho(t,w):= \frac{\sum\limits_{i\in I}\alpha_t^i (F_i(w)-F_i^\star)}{F(w)-\sum\limits_{i\in I} p_i F_i^\star}
\end{equation}
and introduce the following notations:
\begin{equation}
    \overline{\rho} := \min\limits_{t = 0 \hspace{-0.5em} \mod E} ~\rho(t,w_t)
\quad\mbox{and}\quad
    \Tilde{\rho} := \underset{t = 0 \hspace{-0.5em} \mod E}{\max}~ \rho(t,w^\star)
\end{equation}
\begin{remark}\label{defrho}
    We observe that $\rho(t,w)$ is not defined when $F(w)=\sum\limits_{i\in I} p_i F_i^\star$. From now on, it will be implicit that we are not in this case.
\end{remark}
\subsection{Main theorem and corollary}
This theorem is an adaptation of \cite{Cho2020ClientSI}'s main theorem in our framework.
\begin{theorem}\label{MainNONIID}
In framework \ref{framework} under assumptions \ref{assumption}, we obtain the following result: 
\begin{flalign}
   \mathds{E}\left[\Abs{w_{t+1}-w^\star}^2\right] & \le \left(1-\eta_t \mu\left(1+\frac{3}{8}\overline{\rho}\right)\right) \mathds{E}\left[\Abs{w_{t}-w^\star}^2\right] \notag {} + \eta_t^2 \left(32 E^2 G^2 + 6 \overline{\rho} L \Gamma + \sigma^2\right) \notag
   {} +  2 \eta_t \Gamma \left(\Tilde{\rho}-\overline{\rho}\right) \notag\\
\end{flalign}
\end{theorem}
\begin{corollary}\label{eq}
From Theorem \ref{MainNONIID} by assuming $\eta_t = \frac{1}{\mu (t+\gamma)}$ and $\gamma = \frac{4L}{\mu}$, we obtain the following bound:
\begin{equation}
\mathds{E}[F(w_T)] - F^\star \le \frac{1}{T+\gamma} V(\overline{\rho}, \Tilde{\rho}) + E(\overline{\rho}, \Tilde{\rho}) 
\end{equation}
where:
\begin{align}
& V(\overline{\rho}, \Tilde{\rho}) = \frac{4L(32\tau^2 G^2 + \sigma^2)}{3\mu^2 \overline{\rho}} + \frac{8L^2 \Gamma}{\mu^2} + \frac{L\gamma \Abs{w^0-w^\star}^2}{2} \notag\\
& E(\overline{\rho}, \Tilde{\rho}) = \frac{8L\Gamma}{3\mu}\left(\frac{\Tilde{\rho}}{\overline{\rho}}-1\right) \notag\\ \notag
\end{align}
\end{corollary}
\begin{remark}
From the previous result, we have :
\begin{equation}
    E[F(w_T)-F^\star] = O(1/T)
\end{equation}
\end{remark}
\begin{remark}
All the theoretical results presented were obtained using SGD. The choice of this algorithm was motivated by practice and applications, we could have used an optimized version of GD to gain better bounds as in \cite{DBLP:journals/corr/abs-2108-04755}.
\end{remark}
\subsection{Algorithmic consequences of the main results}\label{AlgorithmicCon}
The mathematical bounds $V$  and $E$ can be interpreted respectively as convergence speed and  error estimation. Then, an optimization problem arises in establishing a priority in optimizing these two quantities, without underestimating the trade-off between them. Actually, $V$ and $E$  only suggest the global trend and do not allow to state a proof of universal optimality for the strategies presented below. Let us introduce, first, a remark of interest: 
\begin{remark}
    In Corollary \ref{eq}, since $\frac{8L^2 \Gamma}{\mu^2} + \frac{L\gamma \Abs{w^0-w^\star}^2}{2}$ is a constant depending only on the data set and the first guess, and $\overline{\rho}$ can be arbitrary big, we observe how there exists a limit convergence speed $V_{\min} := \frac{8L^2 \Gamma}{\mu^2} + \frac{L\gamma \Abs{w^0-w^\star}^2}{2}$.
\end{remark}
\subsubsection{Error-free case}
We first analyze the case where $\Gamma = 0$, which corresponds to an IID-dataset. Under this assumption, the following approximation for Theorem \ref{MainNONIID} holds:
\begin{equation}
   \mathds{E}\left[\Abs{w_{t+1}-w^\star}^2\right]  \le \left(1-\eta_t \mu\left(1+\frac{3}{8}\overline{\rho}\right)\right) \mathds{E}\left[\Abs{w_{t}-w^\star}^2\right]   + \eta_t^2 \left(32 E^2 G^2 + \sigma^2\right)
\end{equation}
as well as the following for Corollary \ref{eq}: 
\begin{equation}
    \mathds{E}[F(w_T)] - F^\star \le \frac{1}{T+\gamma}\left[\frac{4L(32\tau^2 G^2 + \sigma^2)}{3\mu^2 \overline{\rho}} + \frac{L\gamma \Abs{w^0-w^\star}^2}{2}\right]
\end{equation}
This setting is interesting because the error term vanishes and then the optimal algorithm is given by the maximization of $\overline{\rho}$, achieved by taking 
\[\displaystyle
    \alpha_t^i = \left\{
    \begin{array}{ll}
        \displaystyle \frac{1}{|J_t|} & \mbox{if } i\in J_t \\
        \\
        \displaystyle 0 & \mbox{else } 
    \end{array}
\right.
\]
where $J_t= \underset{i\in I}{\argmax} (F_i(w_t)-F_i^\star)$.

\begin{remark}
$\overline{\rho}$ is well defined as long as $F(w_t)\neq F(w^\star)$ for all $t$, which is a reasonable assumption.
\end{remark}

\subsubsection{General case}
In the general case,  both $V$ and $E$ depend on the choice of the $\alpha_{i}^t$. This arises a multi-objective optimization problem where we can't minimize $V$ and $E$ simultaneously, as evidenced by the extreme strategies presented below. Consequently, we definitely need to take into account the trade-off between convergence speed and accuracy.
\paragraph{Convergence speed maximization} 
Optimizing the convergence speed, while forgetting about the error, amounts to maximize $\overline{\rho}$, exactly as done in the \emph{Error-free} case. We then get the same strategy.

\begin{remark}
It is important to notice how  this strategy is unreliable: in fact  mathematical analysis cannot guarantee a precise bound over the error term. 
\end{remark}
\paragraph{Error minimization} Instead, minimizing $E(\overline{\rho}, \Tilde{\rho})$ neglecting $V$, amounts to minimize $\frac{\Tilde{\rho}}{\overline{\rho}}-1$. This is achieved when $\alpha_t^i = p_i$, which gives $E=0$.
\begin{remark}
Again, as above, there is no mathematical way to assess a bound on the convergence speed, which could be arbitrarily low. 
\end{remark}
\paragraph{General bounds} Now, knowing that $\alpha_t^i=p_i$ ensures optimal accuracy, we assume $\alpha_t^i=\kappa_t^i p_i$. The following notation is used: $\pi_t= \underset{i\in I}{\min}~\kappa_t^i$, $\Pi_t = \underset{i\in I}{\max~} \kappa_t^i$, $\pi = \underset{t}{\min}~\pi_t$ and $\Pi = \underset{t}{\max~} \Pi^t$.\\ 
Without loss of generality, we assume that $\forall t, \pi_t > 0$ (if it is not the case, we assign to the $\alpha_t^i$ equal to zero an infinitesimal value and increment the other  $\alpha_t^i$ substantially). Under these assumptions, we have that $\frac{\Tilde{\rho}}{\overline{\rho}} \le \frac{\Pi}{\pi}$, and $\frac{1}{\overline{\rho}}\le \frac{1}{\pi}$ and then: 
\begin{equation}
  \mathds{E}[F(w_T)]-F^\star \le \frac{1}{T+\gamma}\left[C + \frac{\lambda_1}{\pi} \right] + \lambda_2 (\frac{\Pi - \pi}{\pi})  
\end{equation}  
where $C, \lambda_1$ and $\lambda_2$ are constants.\\
Moreover, we assuming that there is only one client reaching the max and the min loss, imposing $p_i = 1/N$ (obtaining $\kappa_t^i = n\alpha_t^i$), we can infer the following \emph{min-max} bounds:
\begin{equation}
    \Pi_t/N \le 1 - (N-1)\pi_t/N \xrightarrow{}\Pi \le N - (N-1)\pi \xrightarrow{}E(\overline{\rho},\Tilde{\rho})\le \lambda_2 \frac{N(1-\pi)}{\pi}\le \lambda_2 (\frac{N}{\pi}-N)
\end{equation}
from which, we obtain the inequality below allowing us to have a bound on both $V$ and $E$:
\begin{equation}
  \mathds{E}[F(w_T)]-F^\star \le \frac{1}{T+\gamma}\left[C + \frac{\lambda_1}{\pi} \right] + \lambda_2 (\frac{N}{\pi}-N)
\end{equation}  
\begin{remark}
We cannot infer further results because 
terms $C+\frac{\lambda_1}{\pi}$ and $\lambda_2(\frac{N}{\pi}-N)$ are not tight.
\end{remark}
We conclude this section by observing that in the most general case, we have the following bound:
\begin{equation}
    \Pi \min~p_i \le \max~ \kappa_t^i p_i \le 1 -(N-1) \min~ \kappa_t^i p_i\le 1 - (N-1)\pi \min~p_i
\end{equation}
from here, we infer that $\Pi \le \frac{1-(N-1)\pi \min~p_i}{\min~ p_i}$ and $E \le \frac{1}{\pi \min~p_i}-N$, obtaining :
\begin{equation}\label{Finalbound}
  \mathds{E}[F(w_T)]-F^\star \le \frac{1}{T+\gamma}\left[C + \frac{\lambda_1}{\pi} \right] + \lambda_2 (\frac{1}{\pi \min~p_i}-N)
\end{equation}
There is an intrinsic interest in this last inequality allowing to state that the new speed and error bounds depend only on $\pi$ and to ensure a bound on the error term by setting a well chosen minimal value of the $\alpha_t^i$.
\section{Strategies}\label{strategy}
In this section, we list some strategies arising from the above analysis.
\subsection{Pure strategies}\label{theory}
In this subsection, we present the so called pure strategies in opposition to the hybrid strategy introduced in subsection \ref{hybrid}.
\paragraph{FedAvg-generalized} For any $t$ and $i\in I$, we take $\alpha_t^i = p_i$.\\
\emph{Interpretation:}
The first strategy proposed is inspired by \cite{McMahan2017CommunicationEfficientLO}  consisting in considering as  global model the weighted average upon $p_i$ of the local models. 
As observed above, this approach is optimal in terms of accuracy, since $E=0$ and its convergence speed can be bounded as below: 
\begin{equation}
 V = V_{\min} + \frac{4L-32\tau^2 G^2 + \sigma^2}{3\mu^2}
\end{equation}
\paragraph{FedWorse}\label{fedworse}
For any $t$, let us consider the following quantity $J_t= \underset{i\in I}{\argmax} (F_i(w_t)-F_i^\star)$, we take: 
\[\displaystyle
    \alpha_t^i = \left\{
    \begin{array}{ll}
        \displaystyle \frac{1}{|J_t|} & \mbox{if } i\in J_t \\
        \\
        \displaystyle 0 & \mbox{else } 
    \end{array}
\right.
\]
Note that in practice, two distinct clients never have the same value, i.e. $|J_t|=1$.\\
\emph{Interpretation}  This strategy is the first original algorithmic contribution of this paper and  consists in considering as global model the client's local model with the worst performance at the end of the previous communication round. 
This approach partially leverages the difference among the values of the loss functions of the different clients.As observed above this strategy gives an optimal bound on the convergence speed. \vspace{0.15cm}\\
\paragraph{FedWorse($k$)}
This strategy is a variant of the strategy described in the paragraph above, where instead of taking the client with the highest loss, we consider the $m$ clients when sorted by decreasing order of $(F_i(w_t)-F_i^\star)$.
It actually boils down to the client selection strategy Power-of-Choice by \cite{Cho2020ClientSI}.
\\
\paragraph{FedSoftWorse}
For any $t$ and $i\in I$, we take $\alpha_t^i = p_i \exp (T^{-1} (F_i(w_t) - F^\star_i))$ re-normalized.\\
\textit{Interpretation:} This is the softened version of FedWorse. The reason behind the introduction of this method is that reinforces the stability of FedWorse and that it has a theoretical advantage ensuring nonzero values of the $\alpha_i^t$. We can thus apply the bound \ref{Finalbound} obtaining a bound on the error.\vspace{0.2cm}\\
We introduce now the opposites of FedWorse, FedWorse($k$) and FedSoftWorse. This is not justified by theory, but we have considered it following the intuition that in an easy task taking the quickest client's model could enhance the convergence speed.
\paragraph{FedBetter, FedBetter($k$) and FedSoftBetter} 
For any $t$, consider the following quantity $J_t= \underset{i\in I}{\argmin} (F_i(w_t)-F_i^\star)$, we take: 
\[\displaystyle
    \alpha_t^i = \left\{
    \begin{array}{ll}
        \displaystyle \frac{1}{|J_t|} & \mbox{if } i\in J_t \\
        \\
        \displaystyle 0 & \mbox{else } 
    \end{array}
\right.
\]
As above in practice, two clients never have the same value, i.e. $|J_t|=1$.\\
\emph{Interpretation}  This strategy consists in considering as global model the local model of the client with the best performance at the end of the previous communication round. 
This approach partially force the training on the optimization trajectory of the quickest client in learning.\vspace{0.15cm}\\
FedBetter($k$) is defined as the opposite of FedWorse($k$), i.e. instead of taking the $m$ worst clients, we take the $m$ best ones. 
An analogous modification is done for FedSoftBetter. 
\subsection{Hybrid strategies}\label{hybrid}
The previous strategies can be combined with the objective of improving the learning performance. Intuitively FedWorse, FedBetter and the strategies derived from them, when applied to clients with similar loss values, behave badly leading to an early stabilization to an erroneous value. To attenuate this effect, it is reasonable to combine these approaches with traditional ones such as FedAvg.\\
There are two main methods to pursue this combination: the so called  \emph{simulated annealing} (\cite{doi:10.1126/science.220.4598.671}), which gradually mixes the approaches, and the discrete combination consisting in using an approach until a certain fixed value of accuracy and then change it.
\section{Experimental results and analysis}
In this section, we study extensively the behavior of the  aggregation strategies proposed in section \ref{strategy} and we compare them with the work of \cite{McMahan2017CommunicationEfficientLO}. 
In particular, we evaluate the performance of the aggregation strategies in two classification tasks both in the IID and the Non-IID framework in a synthetic distributed setting. It is important to understand that the experimental part should not be considered as a search for a practical confirmation of the theory but rather as an analysis for possible insights that the derived strategies can bring to general classification tasks without the assumptions introduced to prove the main result. 
\subsection{Synthetic datasets} We generate two distinct synthetic datasets, corresponding to the IID and to the non-IID framework. For the first, we  sort data according to labels, choose the cardinality of the different local datasets and distribute the items preserving an identical label distribution over the clients. Instead, for the second, we sort the dataset by label and divide it into multiple contiguous shards according to the criteria given in \cite{McMahan2017CommunicationEfficientLO}. Then, we distribute these shards among our clients generating an unbalanced partition. We do not require clients to have the same number of samples when generating the partitions, however each client has at least one shard. 
\subsection{Tasks description}
We measure the performance of our aggregation strategies among classification task over the datasets MNIST \cite{deng2012mnist} and Fashion-MNIST \cite{https://doi.org/10.48550/arxiv.1708.07747}. 
\subsection{Model}
The model used is a CNN with two $3\times 3$ convolution layers (the first with 32 channels, the second with 64, each followed with $2\times 2$ max pooling), a fully connected layer with 1600 units and ReLu activation, and a final softmax output layer. The local learning algorithm is a simple mini-batch SGD with a batch size fixed at $64$.
\subsection{Hyperparameters}
 We ran all task for a small number of communication rounds (between $100-150$) that is enough to investigate the initial convergence speed and with a total number of clients equal to $100$. The decreasing learning rate is set to $10^{-3}\cdot 0.99^r$ for each communication round $r \in \{ 0, \dots R-1 \}$. The parameter choices follow the standard assumptions (dimension of the batch) and from an extensive empirical selection aimed to maximize the accuracy reached after 100 communication rounds. 
\subsection{Evaluation}
To evaluate the performance of the strategies proposed, we focus our attention on two kinds of measures: the \emph{accuracy} reached after a fixed number of communication rounds (between 100-150) and the $R_{60}$ an index, common in literature, corresponding to the number of communication rounds required to reach a $60\%$ accuracy. We furthermore keep track of the accuracy value and of the loss function at each communication round of the global model. In particular, in the IID context, since in theory the error term is zero, we compare methods on their accuracy after $100$ epochs. Conversely, since in non-IID methods are considered have a non-zero error term, comparisons after $100$ epochs are biased and $R_{60}$ results are a better comparison base.
\subsection{Resources}
All the strategies are implemented within Pytorch and trained on a single NVIDIA Tesla V100 GPU with 16 GB of GPU memory on an AWS instance for circa 160 hours. 
\subsection{Experimental results}\label{experimental}
 In this subsection, we describe the results obtained in the practical part. For sake of clarity, we distinguish the results in three classes at each of which we devote a paragraph.
 \begin{figure}
 \centering
\subfigure{\includegraphics[width=.48\linewidth]{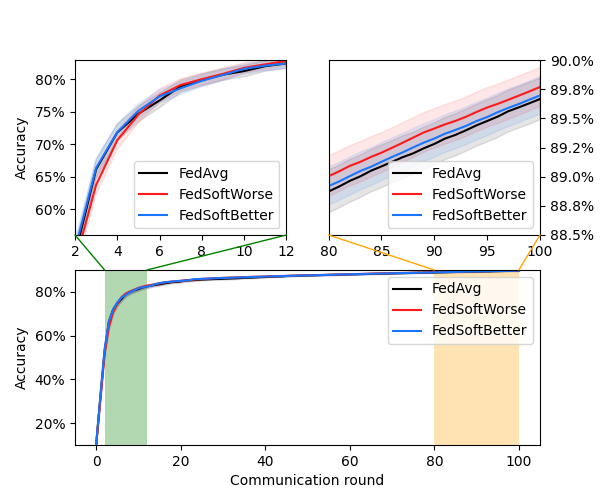}}
\hspace{2mm}
\subfigure{\includegraphics[width=.48\linewidth]{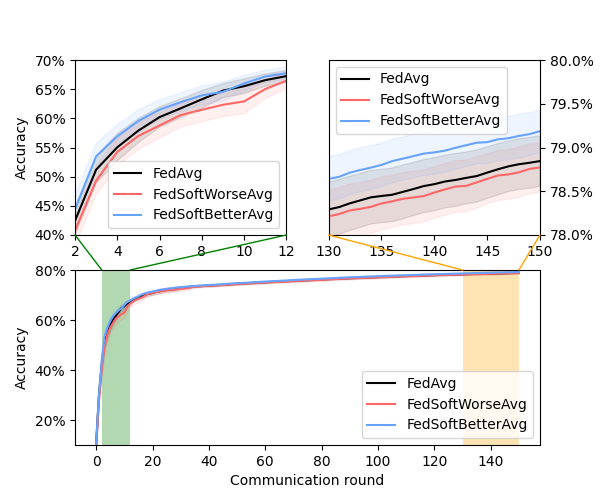}}
\caption{On the left: accuracy curves for MNIST in IID framework with parameters $T=0.2$. On the right: accuracy curves for FMNIST in non-IID framework with hybrid strategies, $T=0.2$ and strategy change at round $20$.}
\end{figure}
\paragraph{Pure strategies in IID framework} We have compared the results in MNIST image classification for the strategies:  FedAvg,  FedSoftBetter and FedSoftWorse.  The results, summarized in Figure 1, are coherent with our theoretical prediction Section \ref{theory}. After 100 global rounds FedSoftWorse has reached the highest accuracy value (89.77\%), followed by FedSoftBetter (89.70\%) and FedAvg (89.67\%). Symmetrically, this result is reflected in the trend of the loss function.\\
 \paragraph{Pure strategies in non-IID framework} 
 We have compared the results in FMNIST image classification for the strategies: FedAvg, FedSoftBetter and FedSoftWorse. In a prior phase, we compared them with their variant without softening but these methods are in practice too instable (Figure \ref{fig:NonIID}, for further observation look at Appendix \ref{non-IID}). The results obtained can be summarized in the table below:  
\vspace{0.15cm}
\begin{center}
\begin{tabular}{ *5c }    \toprule
\emph{Strategies} & \emph{$\textrm{R}_{60}$} \\\midrule
FedAvg & $5.90 \pm 0.37$ \\
FedSoftWorse & $5.73 \pm 0.35$ \\ 
FedSoftBetter & $5.57 \pm  0.32$ \\\bottomrule
 \hline
\end{tabular}  
\end{center}
\vspace{0.15cm}
where we have introduced as error the confidence interval at 95\%. We observe how FedSoftBetter seems to reach with almost half round of advantage the accuracy of 60\%.
We have investigated as well the influence of considering more or less clients in the aggregation step by comparing FedWorse($k$) and FedBetter($k$) for $k=10\%, 20\%$ (more comments in Appendix \ref{kInf}). 
 \paragraph{Hybrid strategies}
These experimental results lead to think that it may be possible to combine some properties of the different strategies do get a better method.
As for the IID framework, in MNIST classification, we get a better performance of FedSoftBetterAvgSoftWorse when compared to pure strategies and other hybrid ones.
As for the non-IID framework, in FMNIST classification FedSoftBetterAvg performs best, which is logic since we combine the initial sprint of FedSoftBetter with the stability of FedAvg.\\
Furthermore, we obtain: 
\vspace{0.15cm}
\begin{center}
\begin{tabular}{ *5c }    \toprule
\emph{Strategies} & \emph{$\textrm{R}_{60}$} \\\midrule
FedAvg & $5.48 \pm 0.75$ \\
FedSoftWorseAvg  &  $6.13 \pm 0.91$ \\ 
FedSoftBetterAvg & $4.97 \pm  0.57$ \\\bottomrule
 \hline
\end{tabular}  
\end{center}
\vspace{0.15cm}
where we have introduced as error the confidence interval at 95\%. The result is clear even though the errors upon the data are significant. More results concerning hybrid strategies are contained in the Appendix \ref{hybPlus}.
\section{Conclusion and discussion}
\paragraph{Contribution} In this work, we have extended the insightful analysis carried out in \cite{Cho2020ClientSI}, and examined further the joint evolution of $\overline{\rho}$ and $\Tilde{\rho}$, giving simpler bounds. From the theoretical results, we have inferred the algorithm FedWorse and from it a few variants with the aim of enhancing stability and utility in practice. We have investigated empirically the hypothetical relaxation of the assumptions and the utility in applications of the strategy derived by evidencing as well the good practice of combining pure strategies together, exploiting the advantage of each.
\paragraph{Limitation}
The results obtained from a theoretical point of view are based on quite stringent assumptions that could be further relaxed.  This relaxation to non-strongly convex case, in particular, seems to be justified, at least empirically; in fact, the experiments are not run in strongly convex framework that seemed to us too restrictive to obtain results useful in real applications. Other weak points of the experimental analysis are the low number of clients (100) that seems far from the real number of nodes in federated learning systems, the dimension and number of the datasets chosen and the difficulty of the classification task: all this quantities should be augmented in further experiments. Furthermore, we were not able to run enough times the experiment to reduce the dimensions of the confidence intervals.
\paragraph{Further work}
Several aspects of interest may be object of further analysis and they are briefly introduced in decreasing order of importance. Both theoretically and practically, it would be useful to analyse the convergence of hybrid strategies obtaining protocols to manage the transition from one method to the other. It would be as well important to study the influence of $\eta$ on the convergence and to find an equivalent of the main theorem when the number of local SGD iterations change. Moreover, it would be interesting to try generalizing our theoretical results to weaker conditions such as Polyak-Łojasiewicz inequality.

\bibliography{bibliography}

\begin{thebibliography}{11}
\providecommand{\natexlab}[1]{#1}
\providecommand{\url}[1]{\texttt{#1}}
\expandafter\ifx\csname urlstyle\endcsname\relax
  \providecommand{\doi}[1]{doi: #1}\else
  \providecommand{\doi}{doi: \begingroup \urlstyle{rm}\Url}\fi

\bibitem[Berahas et~al.(2019)Berahas, Curtis, and
  Zhou]{Berahas2019LimitedmemoryBW}
A.~S. Berahas, F.~E. Curtis, and B.~Zhou.
\newblock Limited-memory bfgs with displacement aggregation.
\newblock \emph{arXiv: Optimization and Control}, 2019.

\bibitem[Cho et~al.(2020)Cho, Wang, and Joshi]{Cho2020ClientSI}
Y.~J. Cho, J.~Wang, and G.~Joshi.
\newblock Client selection in federated learning: Convergence analysis and
  power-of-choice selection strategies.
\newblock \emph{ArXiv}, abs/2010.01243, 2020.

\bibitem[Deng(2012)]{deng2012mnist}
L.~Deng.
\newblock The mnist database of handwritten digit images for machine learning
  research.
\newblock \emph{IEEE Signal Processing Magazine}, 29\penalty0 (6):\penalty0
  141--142, 2012.

\bibitem[Ek et~al.(2021)Ek, Portet, Lalanda, and Vega]{ek:hal-03207411}
S.~Ek, F.~Portet, P.~Lalanda, and G.~Vega.
\newblock {A Federated Learning Aggregation Algorithm for Pervasive Computing:
  Evaluation and Comparison}.
\newblock In \emph{{19th IEEE International Conference on Pervasive Computing
  and Communications PerCom 2021}}, Kassel (virtual), Germany, Mar. 2021.
\newblock URL \url{https://hal.archives-ouvertes.fr/hal-03207411}.

\bibitem[Gower et~al.(2019)Gower, Loizou, Qian, Sailanbayev, Shulgin, and
  Richt{\'a}rik]{Gower2019SGDGA}
R.~M. Gower, N.~Loizou, X.~Qian, A.~Sailanbayev, E.~Shulgin, and
  P.~Richt{\'a}rik.
\newblock Sgd: General analysis and improved rates.
\newblock \emph{ArXiv}, abs/1901.09401, 2019.

\bibitem[Kirkpatrick et~al.(1983)Kirkpatrick, Gelatt, and
  Vecchi]{doi:10.1126/science.220.4598.671}
S.~Kirkpatrick, C.~D. Gelatt, and M.~P. Vecchi.
\newblock Optimization by simulated annealing.
\newblock \emph{Science}, 220\penalty0 (4598):\penalty0 671--680, 1983.
\newblock \doi{10.1126/science.220.4598.671}.
\newblock URL
  \url{https://www.science.org/doi/abs/10.1126/science.220.4598.671}.

\bibitem[Li et~al.(2020)Li, Sahu, Talwalkar, and Smith]{Li2020FederatedLC}
T.~Li, A.~K. Sahu, A.~S. Talwalkar, and V.~Smith.
\newblock Federated learning: Challenges, methods, and future directions.
\newblock \emph{IEEE Signal Processing Magazine}, 37:\penalty0 50--60, 2020.

\bibitem[McMahan et~al.(2017)McMahan, Moore, Ramage, Hampson, and
  y~Arcas]{McMahan2017CommunicationEfficientLO}
H.~B. McMahan, E.~Moore, D.~Ramage, S.~Hampson, and B.~A. y~Arcas.
\newblock Communication-efficient learning of deep networks from decentralized
  data.
\newblock In \emph{AISTATS}, 2017.

\bibitem[Xiao et~al.(2017)Xiao, Rasul, and
  Vollgraf]{https://doi.org/10.48550/arxiv.1708.07747}
H.~Xiao, K.~Rasul, and R.~Vollgraf.
\newblock Fashion-mnist: a novel image dataset for benchmarking machine
  learning algorithms, 2017.
\newblock URL \url{https://arxiv.org/abs/1708.07747}.

\bibitem[Yurochkin et~al.(2019)Yurochkin, Agarwal, Ghosh, Greenewald, Hoang,
  and Khazaeni]{Yurochkin2019BayesianNF}
M.~Yurochkin, M.~Agarwal, S.~S. Ghosh, K.~H. Greenewald, T.~N. Hoang, and
  Y.~Khazaeni.
\newblock Bayesian nonparametric federated learning of neural networks.
\newblock In \emph{ICML}, 2019.

\bibitem[Zhao et~al.(2021)Zhao, Li, and
  Richt{\'{a}}rik]{DBLP:journals/corr/abs-2108-04755}
H.~Zhao, Z.~Li, and P.~Richt{\'{a}}rik.
\newblock Fedpage: {A} fast local stochastic gradient method for
  communication-efficient federated learning.
\newblock \emph{CoRR}, abs/2108.04755, 2021.
\newblock URL \url{https://arxiv.org/abs/2108.04755}.

\end{thebibliography}
\newpage
\section*{Checklist}


\begin{enumerate}
\item For all authors...
\begin{enumerate}
  \item Do the main claims made in the abstract and introduction accurately reflect the paper's contributions and scope?
    \answerYes{The abstract and the introduction discuss the state of the art and clarify the contribution of this work.}
  \item Did you describe the limitations of your work?
    \answerYes{We described extensively the limitation of our work in section 5}
  \item Did you discuss any potential negative societal impacts of your work?
    \answerNA{}
  \item Have you read the ethics review guidelines and ensured that your paper conforms to them?
    \answerYes{}
\end{enumerate}
\item If you are including theoretical results...
\begin{enumerate}
  \item Did you state the full set of assumptions of all theoretical results?
    \answerYes{} 
        \item Did you include complete proofs of all theoretical results?
    \answerYes{}
\end{enumerate}

\item If you ran experiments...
\begin{enumerate}
  \item Did you include the code, data, and instructions needed to reproduce the main experimental results (either in the supplemental material or as a URL)?
    \answerYes{Supplemental material include all you need to  reproduce the experiments}
  \item Did you specify all the training details (e.g., data splits, hyperparameters, how they were chosen)?
    \answerYes{We have explained it briefly in Section 4}
        \item Did you report error bars (e.g., with respect to the random seed after running experiments multiple times)?
    \answerYes{All the results are represented with error bars}
        \item Did you include the total amount of compute and the type of resources used (e.g., type of GPUs, internal cluster, or cloud provider)?
    \answerYes{}
\end{enumerate}

\item If you are using existing assets (e.g., code, data, models) or curating/releasing new assets...
\begin{enumerate}
  \item If your work uses existing assets, did you cite the creators?
    \answerYes{All the existing assets are accurately cited through references in the text.}
  \item Did you mention the license of the assets?
    \answerYes{}
  \item Did you include any new assets either in the supplemental material or as a URL?
    \answerYes{}
  \item Did you discuss whether and how consent was obtained from people whose data you're using/curating?
    \answerNA{}
  \item Did you discuss whether the data you are using/curating contains personally identifiable information or offensive content?
    \answerNA{}
\end{enumerate}
\item If you used crowdsourcing or conducted research with human subjects...
\begin{enumerate}
  \item Did you include the full text of instructions given to participants and screenshots, if applicable?
    \answerNA{We haven't used crowdsourcing or conducted research with human subjects.}
  \item Did you describe any potential participant risks, with links to Institutional Review Board (IRB) approvals, if applicable?
    \answerNA{We haven't used crowdsourcing or conducted research with human subjects.}
  \item Did you include the estimated hourly wage paid to participants and the total amount spent on participant compensation?
    \answerNA{We haven't used crowdsourcing or conducted research with human subjects.}
\end{enumerate}
\end{enumerate}
\newpage
\appendix
\section{Complete version of the proofs}
In this section of the appendix, we report the proofs of the preliminary lemmas and the main results.
The proofs follow nearly identically that of \cite{Cho2020ClientSI} for which we do not claim any novelty credit. However to show how the $\alpha_i^t$ integrate within the proofs, we reproduce here a modified version of the proof of \cite{Cho2020ClientSI}.
\subsection{ Proof of preliminary lemmas}\label{Proof preliminary Lemmas}
\textit{Proof of Lemma \ref{smoothRes}} 
\begin{flalign*}
    f(w)-f(w^\star)-\pscal{\nabla f(w^\star)}{w-w^\star}\geq \frac{1}{2L}\Abs{\nabla f(w)-\nabla f(w^\star)}^2 = \frac{1}{2L}\Abs{\nabla f(w)}^2
\end{flalign*}
where we have used the hypothesis of $L$-smoothness of $f$ and the optimality of $w^\star$.
\begin{flushright}
$\square$
\end{flushright}
\textit{Proof of Lemma \ref{discrepency}}
\begin{flalign*}
    \sum_{i\in I} \alpha_t^i \Abs{w_t-w_t^i}^2 = \sum_{i\in I} \alpha_t^i \Abs{\sum_{j\in I} \alpha_t^j w_t^j -w_t^i}^2
    \le \sum_{i\in I} \alpha_t^i \sum_{j\in I} \alpha_t^j \Abs{w_t^j -w_t^i}^2
    = \sum_{i,j\in I} \alpha_t^i \alpha_t^j \Abs{w_t^j -w_t^i}^2
\end{flalign*}
we have used the definition of $w_t$, Jensen's inequality and now by applying the definitions,  we obtain: 
\begin{flalign*}
    \le \sum_{i,j\in I} \alpha_t^i \alpha_t^j \Abs{\sum\limits_{t=t_0}^{t-1} \eta_t (g_j(w_t^j) - g_i(w_t^i))}^2
\end{flalign*}
where $t_0=\lfloor \frac{t}{E} \rfloor E$
\begin{flalign*}\label{eqJensen}
    \le \eta_{t_0}^2 E \sum_{i,j\in I} \sum\limits_{t=t_0}^{t-1} \alpha_t^i \alpha_t^j \Abs{g_j(w_t^j) - g_i(w_t^i)}^2
    \le \eta_t^2 E \sum_{i,j\in I} \alpha_t^i \alpha_t^j [2\Abs{g_j(w_t^j)}^2+ 2\Abs{g_i(w_t^i)}^2]
\end{flalign*}
by Jensen's inequality.
So applying the expectation and exploiting its linearity, we get: 
\begin{flalign*}
    \mathbb{E}[\sum_{i\in I} \alpha_t^i \Abs{w_t-w_t^i}^2] &\le \eta_t^2 E \  \mathds{E}[\sum_{i,j\in I} \sum\limits_{t=t_0}^{t-1} [2\Abs{g_j(w_t^j)}^2+ 2\Abs{g_i(w_t^i)}^2]]\\
    &\le \eta_{t_0}^2 E \sum_{i,j\in I} \sum\limits_{t=t_0}^{t-1} \alpha_t^i \alpha_t^j 4 G^2
    \le 4 \eta_{t_0}^2 E^2 G^2 \sum_{i,j\in I} \alpha_t^i \alpha_t^j\\
    \intertext{where we used Assumption \ref{assumption4}, then:}
    &\le 16 \eta_t^2 E^2 G^2 \sum_{i,j\in I} \alpha_t^i \alpha_t^j
\end{flalign*}
Since $\sum_{i,j\in I} \alpha_t^i \alpha_t^j = 1$, we get :
\begin{flalign*}
    \sum_{i\in I} \alpha_t^i \Abs{w_t-w_t^i}^2\le 16\eta_t^2 E^2 G^2
\end{flalign*}
\begin{flushright}
$\square$
\end{flushright}
\subsection{ Proof of the main result (Theorem \ref{MainNONIID})}
\begin{equation*}
    \Abs{w_{t+1}-w^\star}^2 = \Abs{w_t - \eta_t \sum\limits_{i\in I} \alpha_t^i g_i(w_t^i) - w^\star}^2
\end{equation*}
\begin{equation*}
    = \Abs{w_t - \eta_t \sum\limits_{i\in I} \alpha_t^i g_i(w_t^i) - w^\star - \eta_t \sum\limits_{i\in I} \alpha_t^i \nabla F_i(w_t^i)+ \eta_t \sum\limits_{i\in I} \alpha_t^i \nabla F_i(w_t^i)}^2
\end{equation*}
where we have used the definition of $w_{t+1}$ and added and subtracted the same term. Now, by developing the square, we have: 
\begin{flalign*}
    &= \Abs{w_t - w^\star - \eta_t \sum\limits_{i\in I} \alpha_t^i \nabla F_i(w_t^i)}^2 + \eta_t^2 \Abs{\sum\limits_{i\in I} \alpha_t^i \nabla F_i(w_t^i) - \sum\limits_{i\in I} \alpha_t^i g_i(w_t^i)}^2 \\
    &\qquad + 2 \eta_t \pscal{w_t - w^\star - \eta_t \sum\limits_{i\in I} \alpha_t^i \nabla F_i(w_t^i)}{\sum\limits_{i\in I} \alpha_t^i \nabla F_i(w_t^i) - \sum\limits_{i\in I} \alpha_t^i g_i(w_t^i)} 
\end{flalign*}
\begin{flalign*}
    &= \Abs{w_t - w^\star}^2 - 2 \eta_t \pscal{w_t - w^\star}{\sum\limits_{i\in I} \alpha_t^i \nabla F_i(w_t^i)} \\
    &\qquad + 2 \eta_t \pscal{w_t - w^\star - \eta_t \sum\limits_{i\in I} \alpha_t^i \nabla F_i(w_t^i)}{\sum\limits_{i\in I} \alpha_t^i \nabla F_i(w_t^i) - \sum\limits_{i\in I} \alpha_t^i g_i(w_t^i)}\\
    &\qquad + \Abs{\eta_t \sum\limits_{i\in I} \alpha_t^i \nabla F_i(w_t^i)}^2 + \eta_t^2 \Abs{\sum\limits_{i\in I} \alpha_t^i \nabla F_i(w_t^i) - \sum\limits_{i\in I} \alpha_t^i g_i(w_t^i)}^2
\end{flalign*}
To compact the notation, we rename the addends as follows: 
\begin{equation*}
    = \Abs{w_t-w^\star}^2 + A_1 +A_2 + A_3 + A_4
\end{equation*}

We will now bound one by one the addends:
\begin{flalign*}
\intertext{\textbf{Bound for  $A_1$}}
    A_1 &= - 2 \eta_t \pscal{w_t - w^\star}{\sum\limits_{i\in I} \alpha_t^i \nabla F_i(w_t^i)}=- 2 \eta_t \sum\limits_{i\in I} \alpha_t^i \pscal{w_t - w^\star}{\nabla F_i(w_t^i)}\\
    &= - 2 \eta_t \sum\limits_{i\in I} \alpha_t^i \pscal{w_t - w_t^i}{\nabla F_i(w_t^i)} - 2 \eta_t \sum\limits_{i\in I} \alpha_t^i \pscal{w_t^i - w^\star}{\nabla F_i(w_t^i)}\\
\intertext{By applying Cauchy-Schwartz and exploiting the convexity of the functions, we have:}
   &\le \eta_t \sum\limits_{i\in I} \alpha_t^i (\frac{1}{\eta_t} \Abs{w_t - w_t^i}^2 + \eta_t \Abs{\nabla F_i(w_t^i)}^2) - 2 \eta_t \sum\limits_{i\in I} \alpha_t^i \pscal{w_t^i - w^\star}{\nabla F_i(w_t^i)}\\
    &= \sum\limits_{i\in I} \alpha_t^i \Abs{w_t - w_t^i}^2 + \eta_t^2 \sum\limits_{i\in I} \alpha_t^i \Abs{\nabla F_i(w_t^i)}^2 - 2 \eta_t \sum\limits_{i\in I} \alpha_t^i \pscal{w_t^i - w^\star}{\nabla F_i(w_t^i)}
\intertext{We then use Assumption \ref{assumption1}, taking advantage from the fact that $\nabla F_i(w_i^\star)=0$:}
    &\le \sum\limits_{i\in I} \alpha_t^i \Abs{w_t - w_t^i}^2 + 2L \eta_t^2 \sum\limits_{i\in I} \alpha_t^i (F_i(w_t^i)-F_i^\star) - 2 \eta_t \sum\limits_{i\in I} \alpha_t^i \pscal{w_t^i - w^\star}{\nabla F_i(w_t^i)}
\intertext{and then by using Assumption \ref{assumption2}, we have:}
    &\le \sum\limits_{i\in I} \alpha_t^i \Abs{w_t - w_t^i}^2 + 2L \eta_t^2 \sum\limits_{i\in I} \alpha_t^i (F_i(w_t^i)-F_i^\star) - 2 \eta_t \sum\limits_{i\in I} \alpha_t^i (F_i(w_t^i)-F_i(w^\star) + \frac{\mu}{2} \Abs{w_t^i-w^\star}^2)
\intertext{Applying Lemma \ref{discrepency} gives the final bound:}
    A_1 &\le 16\eta_t^2 E^2 G^2 - \eta_t \mu \sum\limits_{i\in I} \alpha_t^i\Abs{w_t^i-w^\star}^2 + 2L \eta_t^2 \sum\limits_{i\in I} \alpha_t^i (F_i(w_t^i)-F_i^\star) - 2 \eta_t \sum\limits_{i\in I} \alpha_t^i (F_i(w_t^i)-F_i(w^\star))
\intertext{\textbf{Bound for  $A_2$}}
      A_2&=2 \eta_t \pscal{w_t - w^\star - \eta_t \sum\limits_{i\in I} \alpha_t^i \nabla F_i(w_t^i)}{\sum\limits_{i\in I} \alpha_t^i \nabla F_i(w_t^i)}
\intertext{we observe that its expected value is zero, i.e. $\mathds{E}(A_2)=0$, because of the unbiasedness of the gradient estimator.}
\intertext{\textbf{Bound for $A_3$}}
\intertext{Concerning $A_3$, we only need Assumption \ref{assumption1}:}
    A_3 &= \Abs{\eta_t \sum\limits_{i\in I} \alpha_t^i \nabla F_i(w_t^i)}^2
    \le \eta_t^2 \sum\limits_{i\in I} \alpha_t^i \Abs{\nabla F_i(w_t^i)}^2
    \le 2L\eta_t^2 \sum\limits_{i\in I} \alpha_t^i (F_i(w_t^i)-F_i^\star)
\intertext{\textbf{Bound for $A_4$}}
\intertext{For this bound, we applied  Jensen's inequality and the linearity of the expected value and then Assumption \ref{assumption3}:}
    A_4 &= \mathbb{E}[\eta_t^2 \Abs{\sum\limits_{i\in I} \alpha_t^i \nabla F_i(w_t^i) - \sum\limits_{i\in I} \alpha_t^i g_i(w_t^i)}^2] \le \eta_t^2 \sum\limits_{i\in I} \alpha_t^i \mathbb{E} [\Abs{ \nabla F_i(w_t^i) - \sum\limits_{i\in I} g_i(w_t^i)}^2]\\
    &\le \eta_t^2 \sum\limits_{i\in I} \alpha_t^i \sigma^2 \le \eta_t^2 \sigma^2\\
\intertext{So we get :}
    \mathbb{E}[\Abs{w_{t+1}-w^\star}^2]& \le \mathbb{E}[\Abs{w_t-w^\star}^2] + 16\eta_t^2 E^2 G^2 - \eta_t \mu \mathbb{E}[\sum\limits_{i\in I} \alpha_t^i\Abs{w_t^i-w^\star}^2] + 2L \eta_t^2 \mathbb{E}[\sum\limits_{i\in I} \alpha_t^i (F_i(w_t^i)-F_i^\star)] \\
    &\qquad - 2 \eta_t \mathbb{E}[\sum\limits_{i\in I} \alpha_t^i (F_i(w_t^i)-F_i(w^\star))] + 2L\eta_t^2 \mathbb{E}[\sum\limits_{i\in I} \alpha_t^i (F_i(w_t^i)-F_i^\star)] + \eta_t^2 \sigma^2\\
    &\le (1-\eta_t \mu) \mathbb{E}[\Abs{w_t-w^\star}^2] + 16\eta_t^2 E^2 G^2 + \eta_t^2 \sigma^2 \\
    &\qquad + 4 L \eta_t^2 \mathbb{E}[\sum\limits_{i\in I} \alpha_t^i (F_i(w_t^i)-F_i^\star)] - 2 \eta_t \mathbb{E}[\sum\limits_{i\in I} \alpha_t^i (F_i(w_t^i)-F_i(w^\star))]\\
    &\le (1-\eta_t \mu) \mathbb{E}[\Abs{w_t-w^\star}^2] + 16\eta_t^2 E^2 G^2 + \eta_t^2 \sigma^2 + A_5
\intertext{\textbf{Bound for $A_5$}}
    A_5 &= 4 L \eta_t^2 \mathbb{E}[\sum\limits_{i\in I} \alpha_t^i (F_i(w_t^i)-F_i^\star)] - 2 \eta_t \mathbb{E}[\sum\limits_{i\in I} \alpha_t^i (F_i(w_t^i)-F_i(w^\star))]\\
    &\le 4 L \eta_t^2 \mathbb{E}[\sum\limits_{i\in I} \alpha_t^i F_i(w_t^i)] - 2 \eta_t \mathbb{E}[\sum\limits_{i\in I} \alpha_t^i F_i(w_t^i)] - 2 \eta_t \mathbb{E}[\sum\limits_{i\in I} \alpha_t^i (F_i^\star-F_i(w^\star))] + 2 \eta_t \mathbb{E}[\sum\limits_{i\in I} \alpha_t^i F_i^\star] \\ 
    &\qquad - 4 L \eta_t^2 \mathbb{E}[\sum\limits_{i\in I} \alpha_t^i F_i^\star]\\
    &\le 2\eta_t (2L\eta_t - 1) \mathbb{E}[\sum\limits_{i\in I} \alpha_t^i (F_i(w_t^i)-F_i^\star)] + 2 \eta_t \mathbb{E}[\sum\limits_{i\in I} \alpha_t^i (F_i(w^\star) - F_i^\star)]\\
    &= \mathbb{E}[A_6] + 2 \eta_t \mathbb{E}[\sum\limits_{i\in I} \alpha_t^i (F_i(w^\star) - F_i^\star)]
\intertext{\textbf{Bound for $A_6$}}
\intertext{Let's suppose $\eta_t \le \frac{1}{4L}$. Then :}
    A_6 &= 2\eta_t (2L\eta_t - 1) \sum\limits_{i\in I} \alpha_t^i (F_i(w_t^i)-F_i^\star)\\
    &= 2\eta_t (2L\eta_t - 1) \sum\limits_{i\in I} \alpha_t^i (F_i(w_t^i) - F_i(w_t) + F_i(w_t) - F_i^\star)\\
\intertext{By Assumption \ref{assumption2}, we have :}
    &\le 2\eta_t (1 - 2L\eta_t) \sum\limits_{i\in I} \alpha_t^i [\pscal{\nabla F_i(w_t)}{w_t^i-w_t} + \frac{\mu}{2} \Abs{w_t^i-w_t}^2] - 2\eta_t (2L\eta_t - 1) \sum\limits_{i\in I} \alpha_t^i (F_i(w_t) - F_i^\star)
\intertext{Then Assumption \ref{assumption1} gives :}    
    &\le 2\eta_t (1 - 2L\eta_t) \sum\limits_{i\in I} \alpha_t^i [\eta_t L (F_i(w_t) - F_i^\star) + (\frac{1}{2\eta_t} - \frac{\mu}{2}) \Abs{w_t^i-w_t}^2] \\ 
    &\qquad - 2\eta_t (2L\eta_t - 1) \sum\limits_{i\in I} \alpha_t^i (F_i(w_t) - F_i^\star)\\
    &\le 2\eta_t (2L\eta_t - 1) (1-\eta_t L) \sum\limits_{i\in I} \alpha_t^i (F_i(w_t) - F_i^\star) + (1 - 2L\eta_t)(\frac{1}{2\eta_t} - \frac{\mu}{2}) \sum\limits_{i\in I} \alpha_t^i \Abs{w_t^i-w_t}^2\\
    &\le 2\eta_t (2L\eta_t - 1) (1-\eta_t L) \sum\limits_{i\in I} \alpha_t^i (F_i(w_t) - F_i^\star) + \sum\limits_{i\in I} \alpha_t^i \Abs{w_t^i-w_t}^2
\intertext{So :}
    A_5 &\le \mathbb{E}[A_6] + 2 \eta_t \mathbb{E}[\sum\limits_{i\in I} \alpha_t^i (F_i(w^\star) - F_i^\star)]\\
    &\le 2\eta_t (2L\eta_t - 1) (1-\eta_t L) \mathbb{E}[\sum\limits_{i\in I} \alpha_t^i (F_i(w_t) - F_i^\star)] + \mathbb{E}[\sum\limits_{i\in I} \alpha_t^i \Abs{w_t^i-w_t}^2] \\ 
    &\qquad + 2 \eta_t \mathbb{E}[\sum\limits_{i\in I} \alpha_t^i (F_i(w^\star) - F_i^\star)]\\
\intertext{with Lemma \ref{discrepency} giving}
    &\le 16\eta_t^2 E^2 G^2 + 2\eta_t (2L\eta_t - 1) (1-\eta_t L) \mathbb{E}[\sum\limits_{i\in I} \alpha_t^i (F_i(w_t) - F_i^\star)] + 2 \eta_t \mathbb{E}[\sum\limits_{i\in I} \alpha_t^i (F_i(w^\star) - F_i^\star)]\\
    &= 16\eta_t^2 E^2 G^2 + 2\eta_t (2L\eta_t - 1) (1-\eta_t L) \mathbb{E}[\rho(t, w_t) (F(w_t)-\sum\limits_{i\in I} p_i F_i^\star)] \\ 
    &\qquad + 2 \eta_t \mathbb{E}[\rho(t, w^\star) (F^\star-\sum\limits_{i\in I} p_i F_i^\star)]\\
    &\le 16\eta_t^2 E^2 G^2 + 2\eta_t (2L\eta_t - 1) (1-\eta_t L) \overline{\rho} (\mathbb{E}[F(w_t)]-\sum\limits_{i\in I} p_i F_i^\star) + 2 \eta_t \Tilde{\rho}\Gamma\\
    &= 16\eta_t^2 E^2 G^2 + A_7 + 2 \eta_t \Tilde{\rho}\Gamma
\intertext{\textbf{Bound for $A_7$}}
    A_7 &= 2\eta_t (2L\eta_t - 1) (1-\eta_t L) \overline{\rho} (\mathbb{E}[F(w_t)]-\sum\limits_{i\in I} p_i F_i^\star)\\
    &= 2\eta_t (2L\eta_t - 1) (1-\eta_t L) \overline{\rho} \sum\limits_{i\in I}(\mathbb{E}[F_i(w_t)]- F^\star + F^\star - p_i F_i^\star)\\
    &= 2\eta_t (2L\eta_t - 1) (1-\eta_t L) \overline{\rho} \sum\limits_{i\in I}(\mathbb{E}[F_i(w_t)]- F^\star) + 2\eta_t (2L\eta_t - 1) (1-\eta_t L)\overline{\rho} \sum\limits_{i\in I} (F^\star - p_i F_i^\star)\\
    &= 2\eta_t (2L\eta_t - 1) (1-\eta_t L) \overline{\rho} (\mathbb{E}[F(w_t)]- F^\star) + 2\eta_t (2L\eta_t - 1) (1-\eta_t L) \overline{\rho} \Gamma\\
    &\le 2\eta_t (2L\eta_t - 1) (1-\eta_t L) \frac{\mu}{2} \overline{\rho} \mathbb{E}[\Abs{w_t-w^\star}^2] + 2\eta_t (2L\eta_t - 1) (1-\eta_t L) \overline{\rho} \Gamma\\
    &\le -\frac{3}{8}\eta_t \mu \overline{\rho} \mathbb{E}[\Abs{w_t-w^\star}^2] + 2\eta_t (2L\eta_t - 1) (1-\eta_t L) \overline{\rho} \Gamma\\
    &\le -\frac{3}{8}\eta_t \mu \overline{\rho} \mathbb{E}[\Abs{w_t-w^\star}^2] - 2\eta_t \overline{\rho} \Gamma + 6\eta_t^2 \overline{\rho} L \Gamma
\intertext{So we get:}
    A_5 &\le 16\eta_t^2 E^2 G^2 + A_7 + 2 \eta_t \Tilde{\rho}\Gamma\\
    &\le 16\eta_t^2 E^2 G^2 + \frac{3}{8}\eta_t \mu \overline{\rho} \mathbb{E}[\Abs{w_t-w^\star}^2] - 2\eta_t \overline{\rho} \Gamma + 6\eta_t^2 \overline{\rho} L \Gamma + 2 \eta_t \Tilde{\rho}\Gamma\\
    &\le \eta_t^2 (16 E^2 G^2 + 6 \overline{\rho} L \Gamma) - \frac{3}{8}\eta_t \mu \overline{\rho} \mathbb{E}[\Abs{w_t-w^\star}^2] + 2 \eta_t \Gamma (\Tilde{\rho}-\overline{\rho})
\intertext{\textbf{Main part of the proof:}}
\intertext{Finally we have:}
    \Abs{w_{t+1}-w^\star}^2 &\le (1-\eta_t \mu) \mathbb{E}[\Abs{w_t-w^\star}^2] + 16\eta_t^2 E^2 G^2 + \eta_t^2 \sigma^2 + A_5\\
    &\le (1-\eta_t \mu) \mathbb{E}[\Abs{w_t-w^\star}^2] + 16\eta_t^2 E^2 G^2 + \eta_t^2 \sigma^2 + \eta_t^2 (16 E^2 G^2 + 6 \overline{\rho} L \Gamma)\\ 
    & \qquad - \frac{3}{8}\eta_t \mu \overline{\rho} \mathbb{E}[\Abs{w_t-w^\star}^2] + 2 \eta_t \Gamma (\Tilde{\rho}-\overline{\rho})\\
    &\le (1-\eta_t \mu(1+\frac{3}{8}\overline{\rho})) \mathbb{E}[\Abs{w_t-w^\star}^2] + \eta_t^2 (32 E^2 G^2 + 6 \overline{\rho} L \Gamma + \sigma^2) + 2 \eta_t \Gamma (\Tilde{\rho}-\overline{\rho})
\end{flalign*}
\begin{flushright}
$\square$
\end{flushright}
\subsection{Proof of Corollary \ref{eq}}
We can write Theorem \ref{MainNONIID} as :
\begin{flalign*}
    \Delta_{t+1} \le (1-\eta_t\mu B)\Delta_t + \eta_t^2 C + \eta_t D
\end{flalign*}
with $B = (1+\frac{3}{8}\overline{\rho})$, $C = 32 E^2 G^2 + 6 \overline{\rho} L \Gamma + \sigma^2$ and $D = 2 \Gamma (\Tilde{\rho}-\overline{\rho})$
\\
Let $\psi=\max \left\{\gamma\Abs{w_0-w^\star}^{2}, \frac{1}{\beta \mu B-1}\left(\beta^{2} C+D \beta(t+\gamma)\right)\right\}$,
where $\beta>\frac{1}{\mu B}$, $\gamma>0$.\\
By induction we get that $\forall t, \Delta_{t} \leq \frac{\psi}{t+\gamma}$.\\
Then by the L-smoothness of $F$, we get :
\begin{flalign}
\mathbb{E}\left[F\left(w_t\right)\right]-F^\star \leq \frac{L}{2} \Delta_{t} \leq \frac{L}{2} \frac{\psi}{\gamma+t}
\end{flalign}
\begin{flushright}
$\square$
\end{flushright}
\section{Additional Experimental Results}\label{appB}
\subsection{Some additional results concerning pure strategies in non IID framework}\label{non-IID}
As said in section \ref{experimental}, FedSoftBetter achieves a $60\%$ accuracy faster than FedSoftWorse and FedAvg. However this difference vanishes after some more epochs and FedAvg finally gets better than the two others. 
\begin{figure}[H]
    \centering     \includegraphics[scale=0.4]{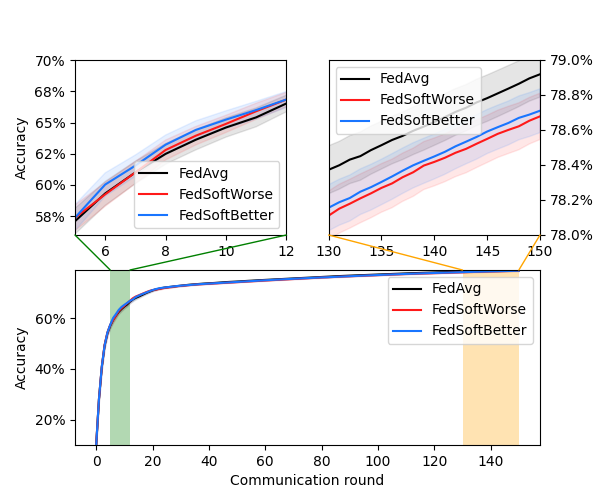}
    \caption{Accuracy curves for FMNIST in non-IID framework with parameter $T=0.2$.}
    \label{fig:NonIID}
\end{figure}
\subsection{Influence of softening procedure on the performance \& influence of increasing the clients' participation in FedBetter and FedWorse}\label{kInf}
We have investigated the influence of the softening procedure upon the efficiency by comparing FedBetter($k$) and FedWorse($k$) with $k=10\%$ of clients and $k=20\%$ of clients upon the task of classifying FMNIST in non IID framework with soften versions (FedSoftBetter and FedSoftWorse). The results are summarized in the figure below, where we can see how softening in fact improves significantly the performance of the methods considered, and how augmenting $k$ from 10\% to 20\% increases the final accuracy.
\begin{figure}[H]
 \centering
\subfigure{\includegraphics[width=.48\linewidth]{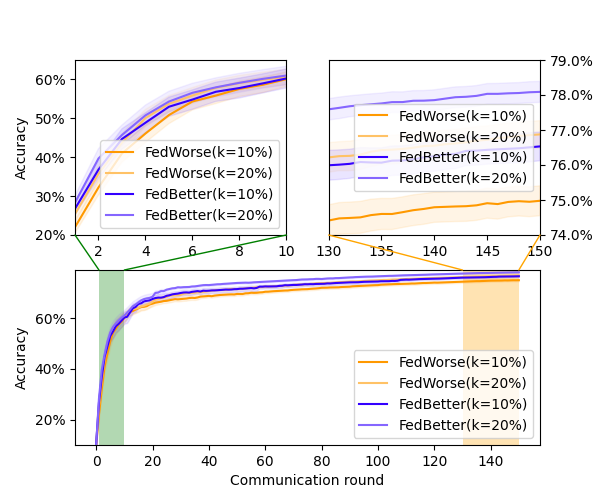}}
\hspace{2mm}
\subfigure{\includegraphics[width=.48\linewidth]{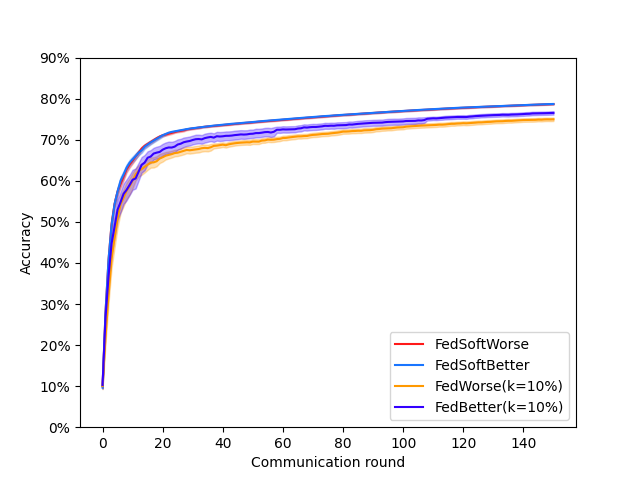}}
\caption{Accuracy curves for FMINST in non-IID framework.}
\end{figure}

\subsection{Some additional results concerning the hybrid methods}\label{hybPlus}
For completeness, we report here the complete results concerning the performance of hybrid methods on FMNIST image classification task in non-IID framework.
\vspace{0.15cm}
\begin{center}
\begin{tabular}{ *5c }    \toprule
\emph{Strategies} & \emph{$\textrm{R}_{60}$} \\\midrule
FedAvg & $ 5.48 \pm 0.75$ \\
FedSoftWorseAvg & $  6.13\pm 0.91$ \\ 
FedSoftBetterAvg & $ 4.97 \pm 0.57$ \\
FedAvgSoftWorse & $ 5.76 \pm  0.72$ \\
FedAvgSoftBetter& $ 5.64 \pm 0.78  $ \\\bottomrule
 \hline
\end{tabular}  
\end{center}
\vspace{0.15cm}

\end{document}